4

# DeepFusionMOT: A 3D Multi-Object Tracking Framework Based on Camera-LiDAR Fusion with Deep Association

Xiyang Wang, Chunyun Fu*, Zhankun Li, Ying Lai, Jiawei He

*Abstract*—In the recent literature, on the one hand, many 3D multi-object tracking (MOT) works have focused on tracking accuracy and neglected computation speed, commonly by designing rather complex cost functions and feature extractors. On the other hand, some methods have focused too much on computation speed at the expense of tracking accuracy. In view of these issues, this paper proposes a robust and fast camera-LiDAR fusion-based MOT method that achieves a good trade-off between accuracy and speed. Relying on the characteristics of camera and LiDAR sensors, an effective deep association mechanism is designed and embedded in the proposed MOT method. This association mechanism realizes tracking of an object in a 2D domain when the object is far away and only detected by the camera, and updating of the 2D trajectory with 3D information obtained when the object appears in the LiDAR field of view to achieve a smooth fusion of 2D and 3D trajectories. Extensive experiments based on the typical datasets indicate that our proposed method presents obvious advantages over the state-of-the-art MOT methods in terms of both tracking accuracy and processing speed. Our code is made publicly available for the benefit of the community[1].

## I. INTRODUCTION

3D multi-object tracking (MOT) is important in various fields such as autonomous driving, security surveillance and mobile robotics. In the existing literature, most MOT methods are devised under a tracking-by-detection framework [1-3], which mainly comprises two steps: 1) object detection, and 2) data association. In recent years, a large number of object detection methods have been proposed, and detection accuracy has been greatly improved. Yet for MOT, challenges still exist in the stage of data association, for instance, dealing with false negatives and false positives due to occlusions.

Data association for MOT has been extensively researched in recent years [4-6]. Sharma et al. [7] proposed to project the trajectory in the current frame to the next frame to find the match directly in the mapping region and calculate corresponding cost function, thereby reducing the search region and the computation cost. Kim et al. [8] proposed a two-stage association mechanism that takes into account characteristics of the camera and LiDAR sensors. It is shown in this work that relatively good tracking results can be achieved by fusing information obtained from both sensors. It should be noted that most existing MOT methods involve merely camera-based 2D/3D tracking or LiDAR-based 3D tracking, and only a few relate to fusion of information obtained from LiDAR and camera sensors for MOT in 3D domain.

In general, a camera can detect a remote object, while this will be difficult for a LiDAR. As a result, for the same trajectory of an object, camera-based methods are able to initialize tracking when the object is far away, but LiDAR-based methods cannot start tracking until the object gets close. The method proposed in this paper achieves good fusion of 2D and 3D trajectories, and tracks an object in a 2D domain once it is detected by a camera and in a 3D domain when it comes within the detection range of a LiDAR sensor. An accurate and reliable four-level deep association mechanism is proposed in this paper, which achieves favourable tracking performance on both KITTI and nuScenes leaderboards. Fig. 1 demonstrates statistics of our method and other competing methods in two key evaluation metrics, Higher Order Tracking Accuracy (HOTA) and Frames Per Second (FPS).

Manuscript received February 23, 2022; Revised May 21, 2022; Accepted June 15, 2022.

This paper was recommended for publication by Editor Eric Marchand upon evaluation of the Associate Editor and Reviewers' comments. This work was supported in part by the National Natural Science Foundation of China under Grant 51805055. (Corresponding author: Chunyun Fu).

This work was supported in part by the National Natural Science Foundation of China under Grant 51805055. (Corresponding author: Chunyun Fu).

Xiyang Wang, Zhankun Li and Jiawei He are with the College of Mechanical and Vehicle Engineering, Chongqing University, Chongqing 400044, China (e-mail: @cqu.edu.cn).

Chunyun Fu is with the State Key Laboratory of Mechanical Transmissions and the College of Mechanical and Vehicle Engineering, Chongqing University, Chongqing 400044, China (e-mail: fuchunyun@cqu.edu.cn).

Ying Lai is with the School of Mechanical Engineering, Guangdong Songshan Polytechnic, Guangdong 512126, China (e-mail: laiying1101@163.com).

[1] https://github.com/wangxiyang2022/DeepFusionMOT

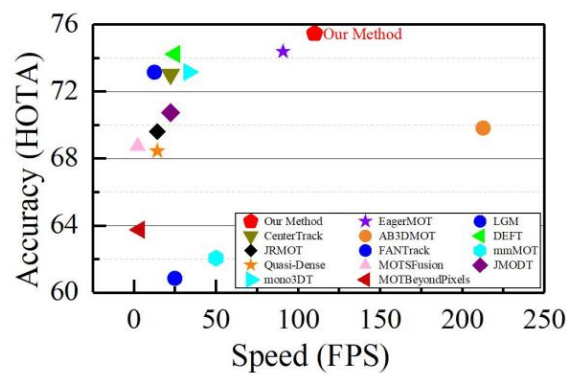

**Fig. 1.** MOT performance of the proposed method on the KITTI 2D MOT leaderboard with relation to several baseline trackers. The higher and the more rightwards, the better. Our method achieves an optimal HOTA and a very high speed.

*A. Literature Review*

*Camera-based Multi-Object Tracking:* Camera-based MOT methods mainly use feature information of an object on RGB images to complete similarity association of the object. The feature information used mainly covers appearance information [9, 10] and motion information [11, 12]. Camera-based MOT methods commonly involve the image domain (2D tracking), yet stereo cameras are also used to realize 3D tracking by extracting depth information. Early camera-based MOT methods (e.g.[2, 13]) follow the tracking-by-detection paradigm. Firstly, an object is detected by a detection algorithm and its trajectory is predicted using a statistical filter (e.g. a Kalman filter). Then, the cost matrix between the predicted trajectories and the detections (e.g. IoU, Euclidean distance, Mahalanobis distance) is calculated. Lastly, data association is formulated as a bipartite graph matching problem that is solved by Hungarian algorithm. Wang et al. [14] and Zhang et al. [15] pointed out that detection and tracking could be performed simultaneously. Specifically, Wang et al. [14] proposed an MOT framework in which target detection and appearance embedding are simultaneously learned in a shared model, which shows an advantage in terms of running time. Zhang et al. [15] solved the bias problem in a network which treats detection as the primary task and tracking as the secondary one, and their approach provides high detection accuracy while maintaining good tracking results. Ballester et al. [16] claimed that adding dynamic object tracking to the front-end of a SLAM system could significantly improve the robustness and accuracy of SLAM in highly dynamic environments.

We see a trend in the recent literature that camera-based MOT methods are combined with other tasks (such as SLAM and instance segmentation), in order to achieve better tracking performance through mutual complementation [16-18]. Nevertheless, camera-based MOT methods are usually 2D tracking solutions which rely on images without depth information. Although some methods have adopted stereo cameras to extract depth information for 3D tracking [19, 20], the accuracy of depth information is inferior to that of LiDAR sensors and the computation load is usually large.

*LiDAR-based Multi-Object Tracking:* LiDAR sensors can provide accurate depth information for 3D tracking. LiDAR-based 3D tracking has started to gain popularity owing to recent breakthroughs in deep learning for point cloud processing [6, 21, 22]. Lately, Weng et al. [23] proposed a simple but fast 3D tracking framework and used only 3D IoU as the cost function for matching. The tracking speed of this approach has reached 207.4 FPS, at the cost of less satisfactory tracking accuracy. Weng et al. [24] proposed a unified 3D MOT and trajectory prediction method. By introducing graphical neural networks, this method improves the discriminative feature learning for MOT and provides contextual information for trajectory prediction, thereby avoiding generation of excessive duplicate trajectory samples. Yin et al. [25] proposed a bounding box center-based framework to represent objects being detected and tracked, thereby converting and simplifying a 3D MOT problem to a nearest point matching problem.

Generally, due to lack of pixel information, the LiDAR-based MOT methods cannot obtain rich appearance information of the objects. Besides, the characteristics of LiDAR sensors render it difficult to detect distant objects, making tracking of distant objects rather difficult if no other sensors are jointly used.

*Fusion-based Multi-Object Tracking:* The fusion of information obtained by camera and LiDAR sensors can compensate for limitations of a single sensor, providing abundant and versatile information on appearance and motion of objects. Huang and Hao [26] proposed a camera-LiDAR fusion-based tracking framework which integrates detection and tracking. In this approach, images and point cloud data are separately sent to a region proposal network to derive respective region features, and then these features are fused and separately sent to an object detection network and an object correlation network. Data association is completed based on mixed-integer programming, using a cost function which takes into account motion similarity and appearance information. This method effectively realizes simultaneous detection and tracking. Frossard and Urtasun [27] proposed a tracking-by-detection approach which combines point cloud data with RGB image information. In this method, the tracking problem is formulated as an inference problem in a deep structured model, and point cloud and RGB data are processed by a pair of feedforward neural networks to generate detection and matching scores. Kim et al. [8] employed a 3D detector and a 2D detector to detect objects from point clouds and images, respectively. Objects detected by the 3D detector are projected onto the images and fused with objects detected by the 2D detector. Then, a two-stage data association procedure is implemented to match the 3D trajectories with the detections. In this approach, a 3D Kalman filter is employed to estimate the states of 3D trajectories. However, the 2D trajectories are not properly taken care of and insufficient 3D information is available to update the 2D trajectories.

Typically, existing sensor-fusion-based MOT methods are focused on designing complex feature extraction networks to fully fuse camera-based 2D features with LiDAR-based 3D features. However, to achieve full performance, most of these approaches need to run on powerful GPUs, making real-time implementations of these methods very difficult.

*B. Problem Statement*

Based on the above discussions, a summary of common problems existing in the relevant literature is given as follows: 1) Depth information required for 3D tracking is normally lacking in the existing camera-based MOT methods. Although some of the methods have used stereo cameras to acquire range information and in turn realized 3D tracking, the computational load is rather large and the accuracy of depth information is not as high as those of LiDAR sensors. On the other hand, LiDAR-based tracking methods are unable to accurately track distant objects due to lack of pixel information. Most existing camera and LiDAR fusion-based tracking methods are designed with complex feature extractors, as a result, these methods usually need to run on GPUs and cannot be easily implemented for real-time applications. 2) Most methods fail to make full use of visual data and point cloud data in the process of camera-LiDAR fusion. Usually, objects detected by LiDAR-based detectors are projected onto images for information extraction. As a result, for objects in the images that are not detected by the LiDAR sensor, corresponding pixel information is lost.

*C. Original Contributions*

To overcome the above shortcomings, a simple, fast and robust 3D tracking framework based on camera-LiDAR fusion is proposed in this paper. The main contributions of this paper are as follows:

- A camera-LiDAR fusion-based 3D real-time tracking framework is proposed, achieving superior MOT performance on the typical tracking dataset.
- A novel deep association mechanism which makes full use of the characteristics of cameras and LiDARs is proposed. This mechanism does not involve any complex cost functions or feature extraction networks, while effectively fusing the 2D and 3D trajectories.
- The proposed tracking framework presents fast computation speed and can be readily implemented in real time.
- The proposed tracking framework can be used in conjunction with arbitrary 2D and 3D detectors, which makes it widely applicable to various scenarios without additional training.

*D. Outline of the Paper*

The rest of the paper is organized as follows. Section II introduces the architecture of the proposed tracking framework. Section III elaborates on the details of the proposed method. Section IV demonstrates the results of comparison with the state-of-the-art tracking solutions. Section V concludes the paper.

## II. SYSTEM ARCHITECTURE

The structure of the proposed camera-LiDAR fusion-based MOT method, as shown in Fig. 2, comprises three main parts: inputs, deep association, and outputs.

For the first part (i.e. inputs), a camera-based 2D detector and a LiDAR-based 3D detector are used to obtain position and motion information of objects in the image domain and the LiDAR domain, respectively. Positions obtained from the LiDAR domain are projected onto the image domain through coordinate transformation, in other words, the 3D bounding boxes in the LiDAR domain are transformed to the 2D bounding boxes in the image domain. Then, the IoU between the 2D bounding boxes obtained from the projection and those obtained from the camera-based detector is calculated, and this IoU is then compared with a threshold to achieve fusion of information acquired by both sensors.

As shown in Fig. 2, the deep association mechanism - the core of the proposed MOT framework - includes four levels of associations: 1) The objects detected by both LiDAR and camera are given the highest priority, and immediately associated with the existing 3D trajectories. 2) After the 1st level of association is completed, the unmatched 3D trajectories are then associated with the objects detected by LiDAR only. 3) In the 3rd level of association, the objects detected by camera only are associated with the 2D trajectories. 4) The 3D trajectories are projected onto the image domain and fused with the 2D trajectories. This step completes the 4th level of association.

The outputs of the proposed MOT framework include 3D and 2D trajectories of objects being tracked.

## III. PROPOSED METHOD

*A. Camera-LiDAR Fusion*

In our method, the detections acquired by camera and LiDAR are used as inputs to the proposed tracking framework. A 2D detector is employed to extract objects from images, and a 3D detector is adopted to extract objects from point clouds. Specifically, 2D information $D_{2d} = (x_c, y_c, w, h)$ of objects in the image domain is obtained by the 2D detector, where $x_c$ and $y_c$ are the center coordinates of the 2D bounding box and $w$ and $h$ denote the width and height of the 2D bounding box. Besides, 3D information $D_{3d} = (x_c, y_c, z_c, w, h, l, \theta)$ of objects in the LiDAR domain is obtained by the 3D detector, where $x_c$, $y_c$ and $z_c$ represent the center coordinates of the 3D bounding box and $w$, $h$, $l$ and $\theta$ are the width, height, length and heading angle of the 3D bounding box.

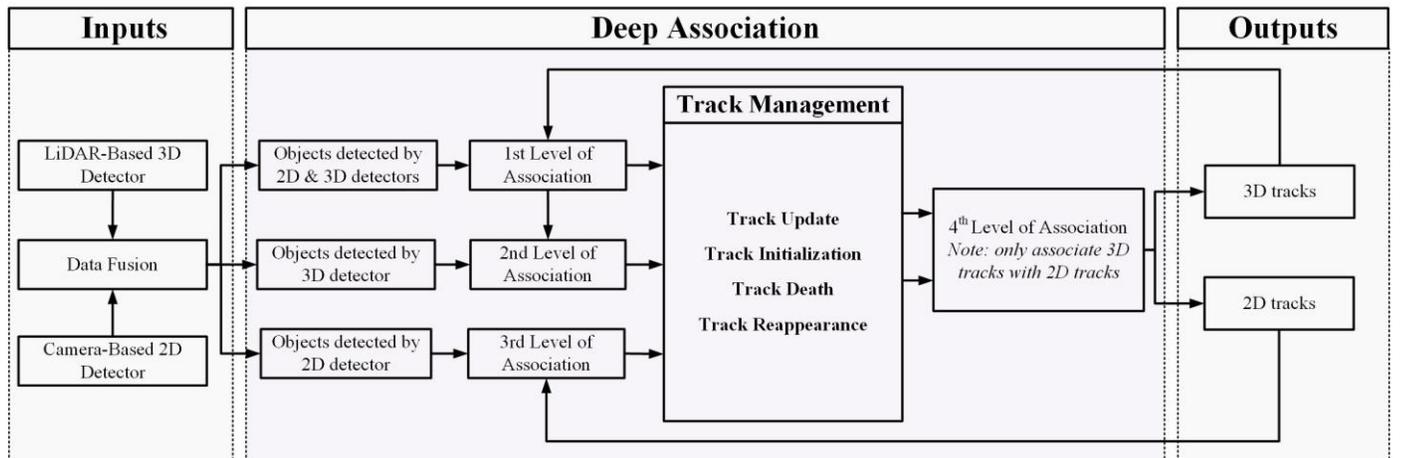

**Fig. 2.** The system architecture of the proposed camera-LiDAR fusion-based method.

Once the two types of bounding boxes are obtained, the 3D bounding box (i.e. $D_{3d}$) is projected onto the image domain through coordinate transformation, which in turn yields a corresponding 2D bounding box (denoted by $D_{2d}^{3d}$ which is different from $D_{2d}$). Then, the IoU between these two types of 2D bounding boxes, $D_{2d}$ and $D_{2d}^{3d}$, is calculated. If this IoU is less than the threshold, we end up with objects in the image domain only (denoted by $D_{2d}^{only}$), as well as objects in the LiDAR domain only (denoted by $D_{3d}^{only}$). If the IoU is greater than the threshold, we obtain objects which exist simultaneously in both domains (denoted by $D_{2d-3d}^{fused}$). Then the foregoing three types of objects are used as inputs to deep association in the next step.

*B. Deep Association*

To make full use of the characteristics of images and point clouds, we propose in this paper a deep association mechanism to tackle the challenging and long-standing data association problem in MOT. This mechanism serves as a good solution to the following three problems: 1) ID switching when an object reappears after being occluded, 2) false negatives and false positives due to missed detections by a detector, and 3) inaccurate tracking of distant objects using LiDAR sensors only. Experiments show that our proposed deep association mechanism works well and outperforms the competing methods when the same 2D and 3D detectors are used. This deep association mechanism comprises four levels of data association, specifically as follows.

*1st Level of Association:* In the 1st level of association, the existing 3D trajectories are associated with the fused detections $D_{2d-3d}^{fused}$, using the following cost function $C_{fused}$:

$$C_{fused} = \begin{cases} C_1^{3d\ iou} & (\text{when } C_1^{3d\ iou} \neq 0) \\ C_2^{dist} & (\text{when } C_1^{3d\ iou} = 0) \end{cases}. \quad (1)$$

$$C_1^{3d\ iou}(i,j) = \frac{d_i \cap t_j}{d_i \cup t_j} \quad (i \in Z^+, j \in Z^+). \quad (2)$$

$$C_2^{dist}(i,j) = \frac{1}{1+\|d_i - t_j\|} \quad (i \in Z^+, j \in Z^+). \quad (3)$$

where $d_i \in D_{2d-3d}^{fused}$ represents the *i*-th detection, $t_j$ denotes the *j*-th trajectory, $C_1^{3d\ iou}(i,j)$ stands for the IoU between the *i*-th detection and the *j*-th trajectory, and $C_2^{dist}(i,j)$ indicates the normalized Euclidean distance between the *i*-th detection and the *j*-th trajectory. It should be noted that in this study, a 2D trajectory is denoted by $T_{2d} = (id, x_c', y_c', w', h')$ and a 3D trajectory is denoted by $T^{3d} = (id, x_c', y_c', z_c', w', h', l', \theta')$, where *id* represents the label or identity of the trajectory.

In equation (1), the fusion of 3D IoU and normalized Euclidean distance is to avoid association failure for fast moving objects. Specifically, when an object reappears after sudden disappearance for a few frames with fast motion, the IoU between the trajectory and the detection can become zero, and as a result the trajectory and the detection cannot be associated. To tackle this issue, Euclidean distance is introduced in the cost function (1) to ensure robust data association for situations where 3D IoU fails. This cost function may lead to three possible association results: matched 3D trajectories (denoted by $T_m^{3d}$), unmatched 3D trajectories (denoted by $T_u^{3d}$), and unmatched 3D detections (denoted by $D_u^{3d}$, representing fused detections that have not been associated with any existing trajectory). The unmatched detections are initialized as new confirmed trajectories, the unmatched trajectories are moved to the 2nd level of data association, and the matched trajectories are updated using their associated detections by means of the updating method proposed in [23].

*2nd Level of Association:* In the 2nd level of association, the unmatched trajectories from the previous stage ($T_u^{3d}$) are associated with the detections only in the LiDAR domain (i.e. $D_{3d}^{only}$). The cost function used in this stage is the same as that in the first stage. If $T_u^{3d}$ can be successfully associated with $D_{3d}^{only}$, then these trajectories are updated using the associated $D_{3d}^{only}$, by means of the updating method proposed in [23]. The remaining unmatched $D_{3d}^{only}$ are initialized as trajectories to be confirmed. It must be noticed that the unmatched trajectories in the second stage differ from those unmatched trajectories in the first stage. Specifically, in this stage, the unmatched trajectories are considered tentative and can be confirmed only if they are successfully associated for three consecutive frames. The underlying reason is that when an object is detected by multiple sensors (e.g. LiDAR and camera), the probability of false detection is quite low. In comparison, there exits a greater chance of false detection if only one of the sensors detects an object at a certain location.

*3rd Level of Association:* The 3rd level of association is only for 2D tracking in the image domain - associating 2D trajectories (denoted by $D_{2d}$) with objects detected by the 2D detector only (i.e. $D_{2d}^{only}$). Generally, cameras can detect objects at a longer distance where LiDAR sensors cannot provide accurate detection. Based on this consideration, a separate 2D tracker is designed in this paper to handle such objects. Once the distance of an object is shortened to fall within the LiDAR field of view (FOV), the 2D trajectory is updated (exactly as in [23]) using the 3D information obtained from the 3D detector to achieve 3D tracking.

*4th Level of Association:* In the 4th level of association, the unmatched 3D trajectories (including the unmatched trajectories in the second stages, and those to be confirmed in the second stage) are associated with trajectories in the image domain in the third stage. Firstly, these 3D trajectories are projected onto the image domain to obtain corresponding 2D bounding boxes. Then, the IoU between each projected 2D bounding box and each originally detected 2D bounding box in the image domain is calculated. Once a 2D trajectory is successfully associated with a 3D trajectory, these two are then fused to form a new 3D trajectory. To achieve fusion, we replace attributes of the 3D trajectory – ID, number of frames of appearance, state of trajectory (confirmed, tentative, dead or reappeared) – with those of the 2D trajectory. Then, we remove the 2D trajectory from the image domain.

The above deep association mechanism has proven to provide superior tracking performance (e.g. low ID switch and fast computation speed) in experiments based on the KITTI and nuScenes tracking datasets.

*C. Track Management*

It is known that a good track management mechanism can well avoid false negatives and false positives [2, 4, 28]. In this paper, the trajectory management approach in [2] is adopted, but the difference is that a new trajectory state - reappeared - is added. Specifically, when a confirmed trajectory is occluded and in turn cannot be associated with any detections for several frames, it is then regarded as a reappeared trajectory. If this reappeared trajectory cannot be associated for the following consecutive frames (greater than a certain threshold), it is then considered disappeared in the sensor FOV and this trajectory becomes dead. Hence, in this paper, a trajectory may have four types of states, including dead, tentative, confirmed and reappeared.

## IV. EXPERIMENTS

In this section, comparative experimental results are demonstrated to illustrate the effectiveness of the proposed 3D tracking framework.

*A. Experimental Setup*

The proposed MOT framework is coded in Python, and implemented on a desktop PC equipped with an Intel Core i9 2.80 GHz CPU and a 32 GB RAM.

*Datasets:* This paper uses the KITTI [29] and the nuScenes [30] datasets for testing. The proposed method excels with state-of-the-art performance using not only the training set but also the testing set in the KITTI dataset. Besides, it also provides good tracking results with the nuScenes dataset.

*Baseline Methods:* To clearly demonstrate the effectiveness of the proposed method, it is compared with the state-of-the-art methods currently available in the literature, including BeyondPixels [7], mmMOT [1], FANTrack [28], AB3DMOT [23], JRMOT [3], MOTSFusion [31], GNN3DMOT [5], JMODT [26], Quasi-Dense [32], EagerMOT [8], LGM [12], DEFT [33] and QD-3DT [34].

*Object Detectors:* For fair comparison with state-of-the-art MOT methods, detectors commonly used in the existing literature are employed in the present study for object detection purposes. Specifically, with the KITTI dataset, RRC [35] is used as the 2D detector, and PointRCNN [22] is employed as the 3D detector. These two detectors have been widely employed in relevant literature, such as MOTSFusion [31], BeyondPixels [7], EagerMOT [8], AB3DMOT [23], and GNN3DMOT [5]. With the nuScenes dataset, Cascade RCNN [36, 37] is used as the 2D detector, and CenterPoint [25] is employed as the 3D detector.

*Evaluation Metrics: 2D MOT Evaluation Metrics:* CLEAR [38] is a commonly used method for evaluating 2D MOT performance, and it includes important evaluation metrics such as Multi-Object Tracking Accuracy (MOTA), Multi-Object Tracking Precision (MOTP), and ID Switch (IDSW). Apart from the CLEAR metrics, HOTA [39], a recently proposed MOT evaluation metric, is currently used by the KITTI dataset as one of the main evaluation metrics for tracking performance. As an evaluation metric that unifies detection quality and association quality, HOTA can be decomposed into several sub-metrics, mainly including Detection Accuracy (DetA) and Association Accuracy (AssA). In this paper, the CLEAR and HOTA metrics are employed for 2D MOT performance evaluation.

*3D Evaluation Metrics:* In this study, the 3D tracking evaluation method proposed in [23] is employed in nuScenes datasets. This method consists of two important evaluation metrics, i.e. Averaged Multi-Object Tracking Accuracy (AMOTA) and scaled Accuracy Multi-Object Tracking Accuracy (sAMOTA).

*B. Experimental Results*

*Quantitative Evaluation:* Based on the KITTI testing set (Car Class) and nuScenes testing set (including 7 categories: car, pedestrian, bicycle, motorcycle, bus, trailer, truck), the method proposed in the present study is compared with other state-of-the-art MOT methods on the leaderboards of these two datasets in terms of various performance metrics. The comparison results are demonstrated in Table I and Table II. It should be pointed out that the data shown in Table I were obtained on May 17, 2022 and those in Table II were obtained on November 19, 2021, and the methods marked with '#' in Table II use the same object detectors as the proposed method in this paper (i.e. RRC 2D detector [35] and/or PointRCNN 3D detector [22]).

TABLE I. 3D TRACKING RESULTS BASED ON THE NUSCENES TESTING SET. OUR PROPOSED METHOD ACHIEVES THE HIGHEST AMOTA, MOTA AND RECALL.

| Method | Input | AMOTA (%)↑ | MOTA (%)↑ | Recall (%)↑ | IDSW↓ |
|---|---|---|---|---|---|
| AB3DMOT [23] | 3D | 15 | 15 | 28 | 9027 |
| DEFT [33] | 2D | 18 | 16 | 34 | 6901 |
| QD-3DT [34] | 2D | 22 | 20 | 38 | 6856 |
| StanfordIPRL-TRI [40] | 3D | 55 | 46 | 60 | **950** |
| Our Method | 2D+3D | **63.5** | **53.3** | **69.6** | 1705 |

Table I shows that our proposed method achieves the highest AMOTA (63.5%), MOTA (53.3%) and Recall (69.6%) using the nuScenes dataset, which indicates that the proposed method provides the best overall tracking performance among the competing methods. Table II shows that for the KITTI dataset tested, the proposed method in this paper achieves the highest HOTA (75.46%), the highest AssA (80.06%) and the least IDSW (84) among all twelve competing MOT methods. In terms of the AssA metric which describes the accuracy of association, the proposed method outperforms other methods with a remarkably higher AssA value. This proves that the association strategy proposed in this paper excels in its robustness. Compared with AB3DMOT, the proposed method improves HOTA by 5.65% using the same 3D detector. Besides, compared with the latest EagerMOT, the proposed method improves HOTA and AssA by 1.07% and 5.90%, respectively. Moreover, IDSW resulting from the proposed method is only 84, while that of EagerMOT reaches 239. Although the other two metrics (DetA and MOTA) of our method are not optimal, they are at the average levels among these state-of-the-art methods.

*Qualitative Evaluation:* In this paper, both the testing and training sets from the KITTI dataset are used to qualitatively evaluate the proposed approach. Fig. 3 shows the visualization results of the proposed method and AB3DMOT using sequence 0002. Specifically, the first column shows the detection results of the 3D detector, the second column displays the detection results of the 2D detector, the third column demonstrates the tracking results of AB3DMOT, the fourth column shows the tracking results of the proposed method, and the fifth column displays the ground truth.

TABLE II. Comparison of 3D MOT Results Using the Testing Set of KITTI-Car. Methods Marked With '#' Use the RRC 2D Detector [35] and/or PointRCNN 3D Detector [22]. The Performance Data of the Eleven Baseline Methods are Sourced From http://www.cvlibs.net/datasets/kitti/eval_tracking.php.

| Method | Published (Year) | Input | HOTA (%)↑ | DetA (%)↑ | AssA (%)↑ | MOTA (%)↑ | IDSW↓ | FPS↑ |
|---|---|---|---|---|---|---|---|---|
| BeyondPixels [7] | ICRA (2018) | 2D+3D | 63.75 | 72.87 | 56.40 | 82.68 | 934 | 3 |
| mmMOT [1] | ICCV (2019) | 2D+3D | 62.05 | 72.29 | 54.02 | 82.23 | 733 | 33 |
| FANTrack [28] | IV (2019) | 2D+3D | 60.85 | 64.36 | 58.69 | 75.84 | 743 | 25 |
| AB3DMOT [23] # | IROS (2020) | 3D | 69.81 | 71.06 | 69.06 | 83.84 | 126 | **213** |
| JRMOT [3] | IROS (2020) | 2D+3D | 69.61 | 73.05 | 66.89 | 85.10 | 271 | 14 |
| MOTSFusion [31] # | RA-L (2020) | 2D+3D | 68.74 | 72.19 | 66.16 | 84.24 | 415 | 2 |
| GNN3DMOT [5] # | CVPR (2020) | 2D+3D | ---- | ---- | ---- | 82.40 | 113 | --- |
| JMODT [26] | IROS (2021) | 2D+3D | 70.73 | 73.45 | 68.76 | 85.35 | 350 | 22 |
| Quasi-Dense [32] | CVPR (2021) | 2D | 68.45 | 72.44 | 65.49 | 84.93 | 313 | 14 |
| EagerMOT [8] # | ICRA (2021) | 2D+3D | 74.39 | **75.27** | 74.16 | **87.82** | 239 | 90 |
| LGM [12] | ICCV (2021) | 2D | 73.14 | 74.61 | 72.31 | 87.60 | 448 | 12 |
| Our Method | --- | 2D+3D | **75.46** | 71.54 | **80.06** | 84.64 | **84** | 110 |

As seen in Fig. 3, the vehicle in the red circle starts being detected by the 2D detector since frame 7, and by the 3D detector since frame 33, which is 26 frames later. For most existing MOT methods, such as AB3DMOT, this vehicle can only be correctly tracked from frame 35 onwards. In comparison, the proposed method is able to initialize tracking in the 2D domain when this vehicle is still far away and detected only by the 2D detector. As shown in the fourth column, this vehicle starts being tracked since frame 9 using the proposed approach. Once it enters the detection range of the 3D detector, i.e. since frame 33, the 2D track is updated using the corresponding 3D information to realize a smooth switch from 2D tracking to 3D tracking. The results in Fig. 3 indicate that the proposed method initializes tracking in an early stage when the 2D detector provides object information, in comparison, the compared method starts tracking much later when the vehicle is successfully detected by its 3D detector. The above results reveal a notable superiority of the proposed MOT method due to effective sensor fusion.

For a vehicle reappearing in the sensor FOV after being occluded for multiple consecutive frames, the proposed method is able to provide stable tracking without causing any IDSW. For example, the vehicle with ID 8214 in Fig. 4 starts being completely occluded by another vehicle since frame 65 and remains occluded for the next 6 consecutive frames. Since frame 71, this vehicle is re-detected and tracking is resumed without having any IDSW.

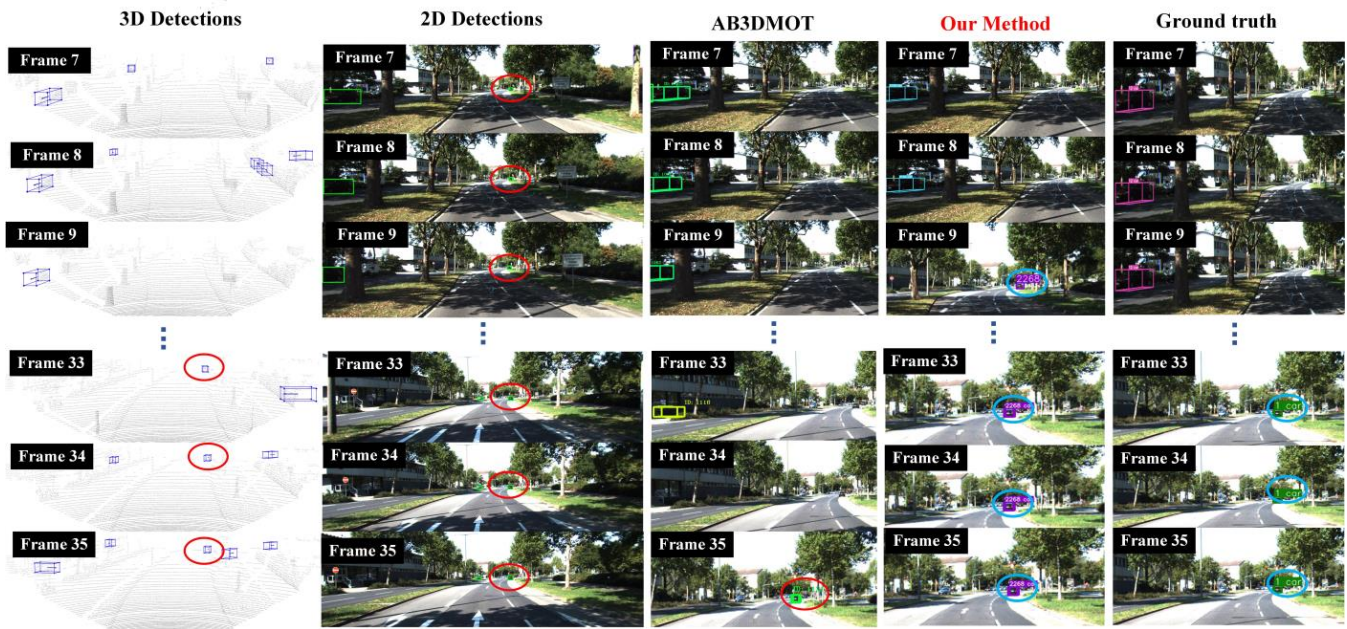

**Fig. 3.** Qualitative Evaluation - An example using sequence 0002 in the validation set where our method outperforms the compared method.

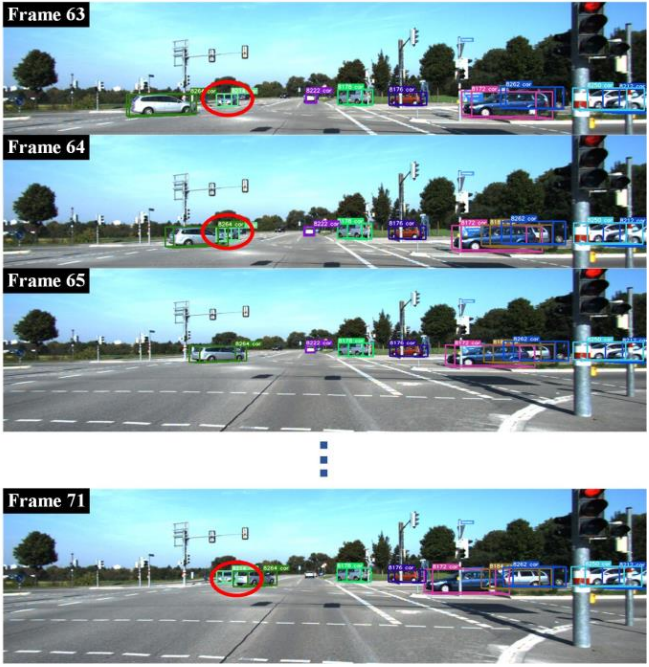

**Fig. 4.** Qualitative Evaluation - An example using sequence 0010 in the testing set where our method maintains the original trajectory without incurring IDSW, when the vehicle is blocked for several consecutive frames and then reappears in the sensor FOV.

*Ablation Study:* In order to investigate the effects of appearance information and object detectors on tracking performance, the KITTI training set (i.e. sequence 0000-0020) is employed in the present study for ablation experiments based on the official KITTI evaluation method [39].

*Effects of Appearance Information:* The feature extractor in [2] is used to extract appearance information and the VeRi dataset [41] is employed for training. As shown in Table III, when appearance information is added, the proposed method delivers slightly higher HOTA and MOTA, with greatly reduced computing speed. The reason for low improvements of tracking performance lies in two aspects: Firstly, the 3D MOT method proposed in this paper is mainly based on point cloud data and the addition of appearance features has little impact on tracking results. Secondly, the feature extractor selected in the ablation experiment is relatively simple, containing only two convolutional layers and six residual blocks. For a trade-off between tracking accuracy and computational speed, the appearance information is not taken into account in the current version of the proposed method. In the next step of studies, a better appearance feature extractor is to be designed based on camera-LiDAR fusion.

TABLE III. ABLATION STUDY - THE RESULTS ARE OBTAINED BY USING DIFFERENT INFORMATION (FOR CAR CLASS).

| Affinity Metrics | HOTA(%)↑ | MOTA(%)↑ | FPS↑ |
|---|---|---|---|
| Euclidean distance + IoU | 77.43 | 87.25 | **104** |
| Appearance + IoU | **77.96** | **88.20** | 12 |

*Effects of Object Detectors:* As mentioned previously, the proposed method can be used in conjunction with arbitrary 2D and 3D detectors. To analyze the impact of different detectors on tracking performance, in this experiment we use PointRCNN [22] and Point-GNN [42] as 3D detectors, and employ RRC [35] and Yolov3 [43] as 2D detectors. Note that for the first three detectors, the detection results provided by their authors are directly employed, while Yolov3 is trained using a server equipped with two NVIDIA GeForce RTX 3090 GPUs. The comparison results are shown in Table IV. It is found that the combination of Point-GNN and Yolov3 delivers the worst tracking performance, and the combination of PointRCNN and Yolov3 provides the fastest tracking performance. Though different choices of detectors can affect tracking results, in general these results do not present significant variations and the tracking performance stays stable.

TABLE IV. ABLATION STUDY - EFFECTS OF USING DIFFERENT OBJECT DETECTORS (FOR CAR CLASS).

| Detector Choice | | HOTA (%)↑ | MOTA (%)↑ | FPS↑ |
|---|---|---|---|---|
| **3D Detector** | **2D Detector** | | | |
| PointRCNN [22] | RRC [35] | **77.45** | **87.28** | 104 |
| PointRCNN [22] | Yolov3 [43] | 77.38 | 86.97 | **148** |
| Point-GNN [42] | RRC [35] | 75.64 | 82.41 | 131 |
| Point-GNN [42] | Yolov3 [43] | 75.15 | 82.18 | 127 |

## V. CONCLUSION AND FUTURE WORK

In this paper, a novel 3D MOT framework based on camera-LiDAR fusion is proposed. In this framework, the characteristics of these two types of sensors are made full use of, and an effective and efficient deep association mechanism which relies only on motion information is proposed for data association. Compared with other state-of-the-art MOT methods, our method achieves the highest HOTA, the highest AssA, and the least IDSW for the dataset tested. Besides, our method can cope well with occlusions while enabling fusion of 2D and 3D trajectories. The ablation experiments indicate that by using only motion information our method remains highly robust, and it can be used in combination with a variety of object detectors, while delivering a fast tracking speed. The proposed MOT framework achieves a good balance between computing speed and tracking accuracy, and is suitable for real-time MOT applications in the field of autonomous driving.

In our next step of research, several aspects of improvement will be made to further enhance the performance of the proposed method. Firstly, LiDAR and visual appearance features, as along with motion features, will be incorporated in the cost function to enhance association robustness. Secondly, inspired by [44], GPS/IMU data will be included to compensate for the prediction errors of targets resulting from the ego-vehicle motions, and appropriate detection filtering and refinement can be added to improve the confidence of detection. Thirdly, based on the concept proposed in [45], more efficient trajectory management mechanisms will be investigated and devised.